\documentclass[runningheads]{llncs}
\usepackage{graphicx}
\usepackage{times}
\usepackage{epsfig}
\usepackage{amsmath}
\usepackage{amssymb}
\usepackage{url}
\usepackage{subfig}
\usepackage{float}
\newcommand{\etal}{\textit{et al.}}
\begin{document}
\title{RN-VID: A Feature Fusion Architecture for Video Object Detection}
%
%
\author{Hughes Perreault\textsuperscript{1}, Maguelonne Heritier\textsuperscript{2}, Pierre Gravel\textsuperscript{2}, Guillaume-Alexandre Bilodeau\textsuperscript{1} and Nicolas Saunier\textsuperscript{1}}
\authorrunning{H. Perreault et al.}
%
\institute{Polytechnique Montreal\textsuperscript{1}, Genetec\textsuperscript{2} \\ \email{\{hughes.perreault, gabilodeau, nicolas.saunier\}@polymtl.ca, \{mheritier, pgravel\}@genetec.ca}}
\maketitle              
\begin{abstract}
Consecutive frames in a video are highly redundant. Therefore, to perform the task of video object detection, executing single frame detectors on every frame without reusing any information is quite wasteful. It is with this idea in mind that we propose RN-VID (standing for RetinaNet-VIDeo), a novel approach to video object detection. Our contributions are twofold. First, we propose a new architecture that allows the usage of information from nearby frames to enhance feature maps. Second, we propose a novel module to merge feature maps of same dimensions using re-ordering of channels and $1 \times 1$ convolutions. We then demonstrate that RN-VID achieves better mean average precision (mAP) than corresponding single frame detectors with little additional cost during inference.

\end{abstract}
\section{Introduction}
Convolutional neural network (CNN) approaches have been dominant in the last few years for solving the task of object detection, and there has been plenty of research in that field. On the other hand, research on video object detection has received a lot less attention. To detect objects in videos, some approaches try to speed up inference by interpolating feature maps~\cite{Liu_2018_CVPR}, while others try to combine feature maps using optical flow warping~\cite{zhu2017flow}. In this work, we present an end-to-end architecture that learns to combine consecutive frames without prior knowledge of motion or temporal relations. 

Even though research on video object detection has been less popular than its single frame counterpart, the applications are not lacking. To name a few: autonomous driving, intelligent traffic systems (ITS), video surveillance, robotics, aeronautics, etc. In today's world, there is a pressing need to build reliable and fast video object detection systems. The number of possible applications will only grow over time. 

\begin{figure}[t]%
    \centering
    \subfloat[]{\includegraphics[width=.16\linewidth]{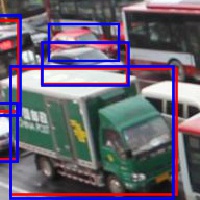}}
    \hspace{2.5em}
    \subfloat[]{\includegraphics[width=.16\linewidth]{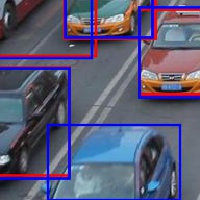}}
    \hspace{2.5em}
    \subfloat[]{\includegraphics[width=.16\linewidth]{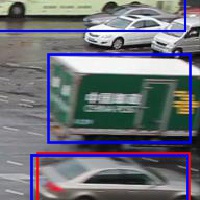}}
    \hspace{2.5em}
    \subfloat[]{\includegraphics[width=.16\linewidth]{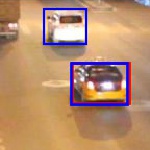}}
    \vspace{-0.5em}
    \caption{Qualitative examples where our model (blue) performs better than the RetinaNet baseline (red). (a) the two cars in the back are heavily occluded by the green truck, (b) the car in the bottom center is being occluded by the frame boundary, (c) the green truck is blurry due to motion blur, (d) as cars become smaller, they become harder to detect, like the white one at the top.}
    \label{challenges}%
\end{figure}

Using multiple frames to detect the objects on a frame presents clear advantages, if used correctly. It can help solve problems like occlusion, motion blur, compression artifacts and small objects (see in figure~\ref{challenges}). When occluded, an object might be difficult or nearly impossible to detect and classify. When moving, or when the camera is moving, motion blur can occur in the image making it more challenging to locate and recognize objects because it changes their appearance. In digital videos, compression artifacts can alter the image quality and make some parts of the frame more difficult to analyze. Small objects can be difficult to locate and recognize, and having multiple frames allows us to use motion information (implicitly or explicitly) as a way to help us find them. Implicitly by letting the network learn how to do it, explicitly by feeding the network optical flow or frame differences. 

Our model relies on the assumption that a neural network can learn to make use of the information in successive frames to address these challenges, and this paper demonstrates the advantages of such a model. Frame after frame, the object instances are repeated several times under slightly different angles, occlusion levels and illuminations, in a way that could be thought as similar to data augmentation techniques. We seek to make the network learn what is the best fusion operation for each feature map channel originating from several frames. Our proposed method contains two main contributions: an object detection architecture based on RetinaNet~\cite{lin2018focal} that merges feature maps of consecutive frames, and a fusion module that merges feature maps without any prior knowledge or handcrafted features. Combined together, these two contributions form an end-to-end trainable framework for video object detection and classification. 

Since this domain contains a lot of interesting challenges and applications, our evaluation is concentrated on traffic surveillance scenes. The effectiveness of our method is evaluated on two popular object detection datasets composed of video sequences, namely UA-DETRAC~\cite{Wen2015Tracking} and UAVDT~\cite{du2018unmanned}. We compare both with the RetinaNet baseline from which we build upon, and state-of-the-art methods from the public benchmarks of those datasets. Results show that our method outperforms both the baseline and the state-of-the art methods.  

\section{Related Work}
\subsection{Object Detection}
Over the last few years, the research focus for object detection has been on single frame detectors. Deep learning-based methods have been dominant on all benchmarks. The two main categories are two-stage detectors, which use a region proposal network, and single-stage detectors, which do not. R-CNN~\cite{rcnn_Girshick_2014_CVPR}, a two-stage detector, was the first dominant object detector to use a CNN. It used an external handcrafted object proposal method called selective search~\cite{uijlings2013selective} to produce bounding boxes. It would then extract features for each bounding box using a CNN and would classify those features using SVM. Fast R-CNN~\cite{Girshick_2015_ICCV_fast} builds upon this idea by addressing the bottleneck (passing each bounding box in a CNN). The way it solves this problem is by computing deep features for the whole image only once and cropping these corresponding features for each bounding box proposals. Faster R-CNN~\cite{ren2015faster} improves furthermore by making the architecture completely trainable end-to-end by using a CNN to produce bounding box proposals, and by performing a classification and regression to refine the proposals. R-FCN~\cite{RFCN_NIPS2016_6465} improves Faster R-CNN by introducing position sensitivity of objects, and by doing so can localize them more precisely. It divides each proposal into a regular grid and classifies each cell separately. In Evolving Boxes~\cite{EB_wang2017evolving}, the authors build an architecture specialized for fast vehicle detection that is composed of a proposal and an early discard sub-network to generate candidates under different feature representation, as well as a fine-tuning sub-network to refine those boxes.  

Single-stage object detectors aim to speed up the inference by removing the object proposal phase. That makes them particularly well suited for real-time applications. The first notable single-stage network to appear was YOLO~\cite{redmon2016you_yolo}, which divides the image into a regular grid and makes each grid cell predict two bounding boxes. The main weakness of YOLO is thus large numbers of small objects, due to the fact that each grid cell can only predict two objects. A high density of small objects is often found in the traffic surveillance context. Two improved versions of YOLO later came out, YOLOv2~\cite{redmon2016you_yolo} and YOLOv3~\cite{redmon2018yolov3}. SSD~\cite{liu2016ssd} tackles the problem of multi-scale detection by combining feature maps at multiple levels and applying a sliding window with anchor boxes at multiple aspect ratio and scale. RetinaNet~\cite{lin2018focal} works similarly to SSD, and introduces a new loss function, called focal loss that addresses the imbalance between foreground and background examples during training. RetinaNet also uses the state-of-the-art way of tackling multi-scale detection, Feature Pyramid Network (FPN)~\cite{Lin_2017_CVPR_FPN}. FPN builds a feature pyramid at multiple levels with the help of lateral and top-down connections and performs classification and regression on each of these levels. 

\subsection{Video Object Detection}
Here we present an overview of some of the most notable work on video object detection. In Flow Guided Feature Aggregation (FGFA)~\cite{zhu2017flow}, the authors use optical flow warping in order to integrate feature maps from temporally close frames, which allows them to increase detection accuracy. In MANet~\cite{wang2018fully}, the authors use a flow estimation and train two networks to perform pixel-level and instance-level calibration. Some works incorporate the temporal aspect explicitly, for example, STMM~\cite{xiao2018video} uses a recurrent neural network to model the motion and the appearance change of an object of interest over time. Other works focus on increasing processing speed by interpolating feature maps of intermediate frames, for instance in~\cite{Liu_2018_CVPR} where convolutional Long Short-Term Memories (LSTMs) are used. These previous works use some kind of handcrafted features (temporal or motion), while our work aims to train a fusion module completely end-to-end. Kim \etal~\cite{Bertasius_2018_ECCV} trained a model by using deformable convolutions that could compute an offset between frames. Doing so allowed them to sample features from close frames to better detect objects in a current frame. This helps them in cases of occlusion or blurriness. In 3D-DETNet~\cite{3D_detnet_li20183d}, to combine several frames, the authors focus on using 3D convolutions on concatenated features maps, generated from consecutive frames, to improve them. MF-SSD~\cite{broad2018recurrent_mf-ssd}, standing for Recurrent Multi-frame Single Shot Detector, extends the SSD~\cite{liu2016ssd} architecture to merge features of multiple sequential frames with a recurrent convolutional module. Perreault \etal~\cite{perreault2019road} trained a network on concatenated image pairs for object detection but could not benefit from pre-trained weights and therefore had to train the network from scratch to outperform the detection on a single frame. 

\subsection{Optical flow by CNNs} 
Works on optical flow by CNNs showed that we can train a network to learn motion from a pair of images. Therefore, similar to our goal, these works put together information from consecutive frames. FlowNet~\cite{flownet_Dosovitskiy_2015_ICCV} is the most notorious work in this field, being the first to present an end-to-end trainable network for estimating optical flow. In the paper, two models are presented, FlowNetSimple and FlowNetCorr. Both models are trained on an artificial dataset of 3D models of chairs. FlowNetSimple consists of a network that takes as input a pair of concatenated images, while FlowNetCorr used a correlation map between higher level representation of each image of the pair. The authors later released an improved version named FlowNet~2.0~\cite{flownet2_Ilg_2017_CVPR} that works by stacking several slightly different versions of FlowNet on top of each other to gradually refine the flow.

\section{Proposed Method}
Formally, the problem we want to solve is as follows: given a target image, a window of $n$ preceding and $n$ future frames and predetermined types of objects, place a bounding box around and classify every object of the predetermined types in the target image.   

To address this problem, we propose two main contributions, a novel architecture for object detection and a fusion module to merge feature maps of the same dimensions. We crafted this architecture to allow the usage of pre-trained weights from ImageNet~\cite{imagenet_cvpr09} in order to build over methods from the state-of-the-art. 

\subsection{Baseline: RetinaNet}
We chose to use the RetinaNet~\cite{lin2018focal} as a baseline upon which to build our model, due to its high speed and good performance. To perform detection at various scales, RetinaNet uses an FPN, which is a pyramid of feature maps at multiple scales (see figure~\ref{architecture}). The pyramid is created with top-down and side connections from the deepest layers in the network, and going back towards the input, thus growing in spatial dimension. A sliding window with boxes created with multiple scales and aspect ratios is then applied at each pyramid level. Afterwards, every box is passed through a classification and a regression sub-network. Finally, non maximal suppression is performed to remove duplicates. The detections with the highest confidence scores are the ones that are kept. As a backbone extractor, we used VGG-16~\cite{VGG_Simonyan2014} for the good trade-off between speed and size that it offers. RetinaNet uses the focal loss for classification:
\begin{equation}
    FL(p')=-\alpha_t(1-p')^\gamma log(p')
\end{equation}

where $\gamma$ is a factor that diminish the loss contributed by easy examples. $\alpha_t$ is the inverse class frequency, and its purpose is to give more representation to underrepresented classes during training. $p'$ is the probability that the predicted label corresponds to the ground-truth label. 

So, if the network predicts with a high probability and is correct, or a low probability and is incorrect, the loss will be marginally affected due to those examples being easy. For the cases where the network is confident (high probability) and incorrect at the same time, the examples will be considered hard and the loss will be affected more.

\subsection{Model Architecture}

\begin{figure*}
\begin{center}
\includegraphics[width=0.8\linewidth]{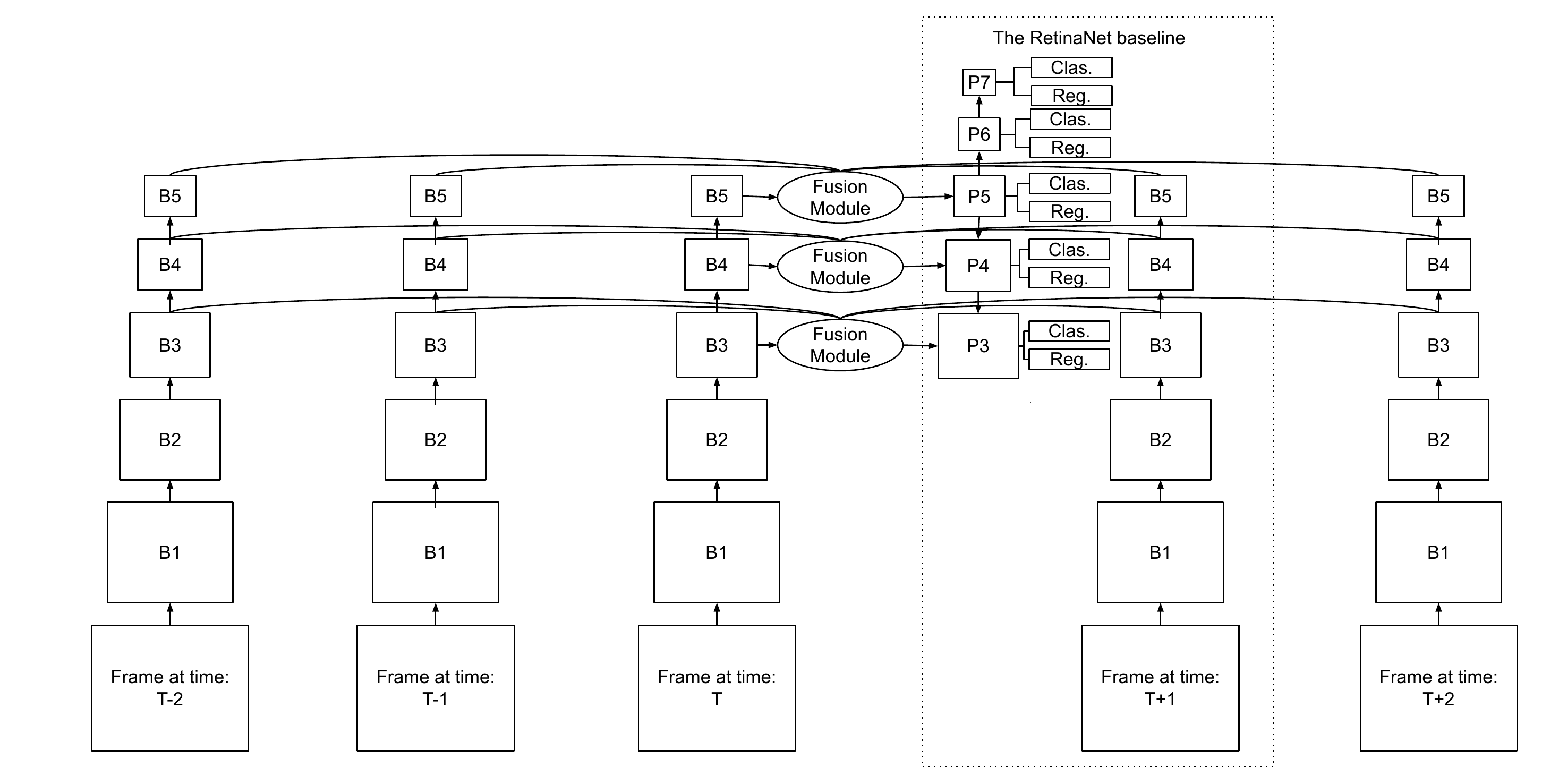}
\end{center}
\vspace{-1em}
   \caption{A representation of our architecture with $n=2$. Each frame is passed through a pre-trained VGG-16, and the outputs of block 3, block 4 and block 5 are collected for fusion. B1 to B5 are the standard VGG-16~\cite{VGG_Simonyan2014} blocks, and P3 to P7 are the feature pyramid levels. In the dotted frame is an overview of our baseline, a RetinaNet~\cite{lin2018focal} with VGG-16 as a backbone.}
\label{architecture}
\end{figure*}

The main idea of the proposed architecture is to be able to compute features for every frame of a sequence only once, and to be able to use these pre-computed features to enhance the features for a target frame t. The fusion module thus comes somewhat late in the network. 

Our network uses multiple input streams that eventually merge into a single output stream, as shown in figure~\ref{architecture}. For computing the feature pyramid for a frame at time $t$, we will use $n$ preceding frames and $n$ future frames. All the $2n + 1$ frames are passed through the VGG-16 network, and we keep the outputs of blocks B3, B4 and B5 for each frame. In RetinaNet, these outputs are used to create the feature pyramid. We then use our fusion module to merge the corresponding feature maps of each frame (block B3 outputs together, block B4 outputs together, etc.) in order to enhance them, before building the feature pyramid. This allows us to have higher quality features and to better localize objects. We then use the enhanced maps as in the original RetinaNet to build the feature pyramid. 

During the training process, we have to use multiple frames for one ground-truth example, thus slowing down the training process. However, for inference on video, the features computed for each frame are used multiple times making the processing time almost identical to the single frame baseline. 

\subsection{Fusion Module}
\begin{figure}[t]
\begin{center}
\includegraphics[width=0.42\linewidth]{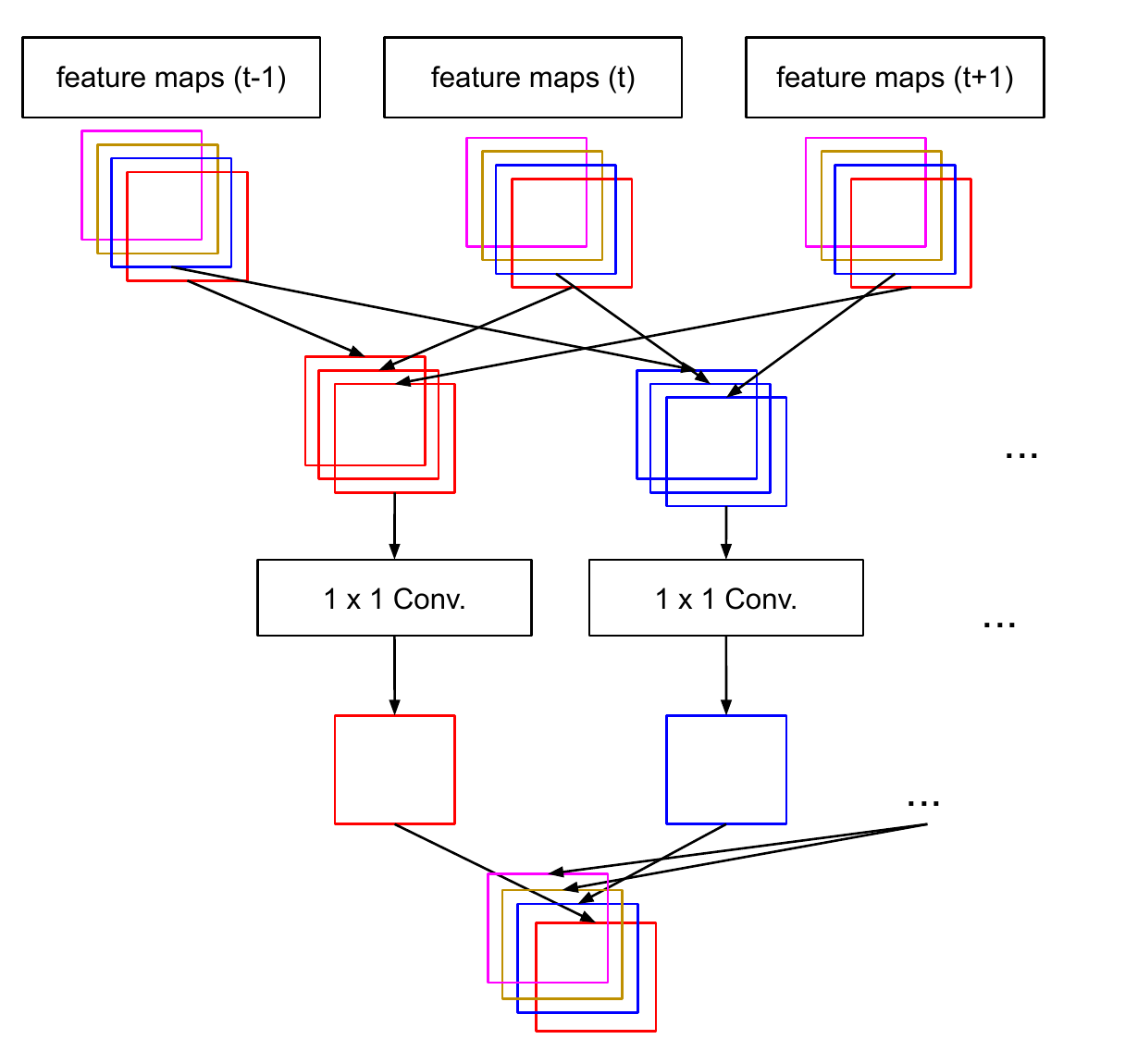}
\vspace{-1em}
\end{center}
   \caption{Our fusion module consists of channel re-ordering, concatenation, $1 \times 1$ convolution, and a final concatenation (better seen in color).}
\label{fusion}
\end{figure}

In order to combine equivalent feature maps of consecutive frames, we designed a lightweight and trainable feature fusion module (see figure~\ref{fusion}). The inspiration for this module is the various possible way a human would do the task. Let us say you wanted to combine feature map channels of multiple consecutive frames. Maybe you would look for the strongest responses and only keep those, making the merge operation an element-wise maximum. Maybe you would want to average the responses over all the frames. This `merge operation' might not be the same for all channels. The idea is to have a network learn the best way to merge feature maps for each channel separately, with $1 \times 1$ convolutions over the channels. 

In our fusion module, we use $1 \times 1$ convolutions in order to reduce the dimension of tensors. In the Inception module~\cite{szegedy2015going} of the GoogLeNet, the $1 \times 1$ convolution is actually used as a way to reduce the dimensions which inspired our work. The inception module allowed them to build a deeper and wider network while staying computationally efficient. In contrast, in our work, we use $1 \times 1$ convolutions for learning to merge feature maps. 

The module takes as input $2n + 1$ feature maps of dimension $w * h * c$ (for width, height and channels respectively), and outputs a single enhanced feature maps of dimension $w * h * c$. The feature maps that we are combining come from corresponding pre-trained VGG-16 layers, so it is reasonable to think that corresponding channels are responses from corresponding `filters'. The idea is to take all the corresponding channels from the consecutive frames, and combine them to end up with only one channel, and thus re-build the wanted feature map, as shown in figure~\ref{fusion}. 

Formally, for $2n + 1$ feature maps of dimension $w * h * c$, we extract each $c$ channels one by one and concatenate them, ending up with $c$ tensors of dimension $w * h * (2n + 1)$. We then perform a 2D convolution with a $1 \times 1$ convolution kernel ($1 * 1 * (2n + 1)$) on the $c$ tensors, getting $c$ times $w * h * 1$ as an output. The final step is to concatenate the tensors channel-wise to finally get the $w * h * c$ tensor that we need. The module is entirely learned, so we can interpret this module as the network learning the operation that best combines feature maps, for each channel specifically, without any prior knowledge or handcrafted features. 

\section{Experiments}

\subsection{Datasets}

\begin{figure}[t]%
    \centering
    \subfloat[]{\includegraphics[height=6em]{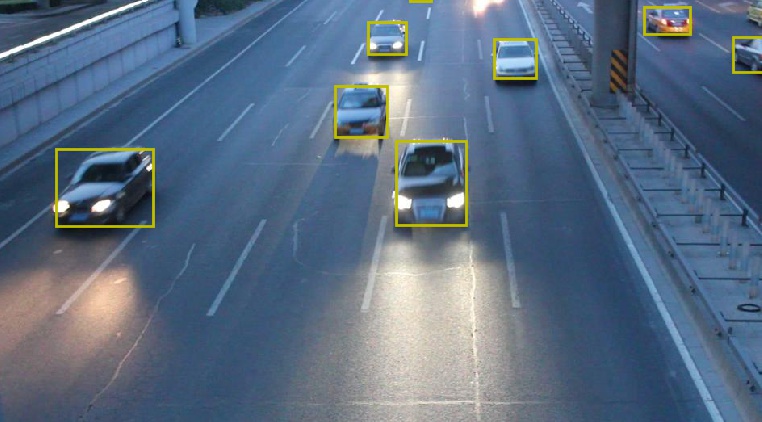}\label{UA-DETRAC}}
    \hspace{3.5em}
    \subfloat[]{\includegraphics[height=6em]{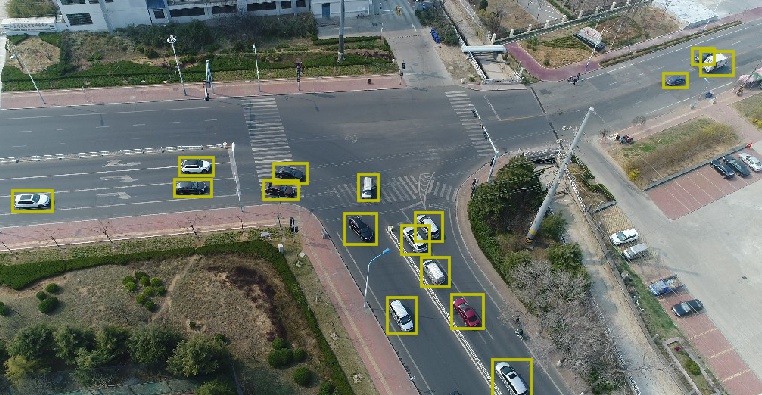}\label{UAVDT}}
    \vspace{-0.5em}
    \caption{(a) An example frame of UA-DETRAC and its ground-truth annotations. (b) An example frame of UAVDT and its ground-truth annotations.}
\end{figure}

The training process of our method requires consecutive images from videos. We chose two datasets containing sequences of moving road users: UA-DETRAC~\cite{Wen2015Tracking} (fixed camera, 960x540, 70000 images, 4 possible labels, see figure~\ref{UA-DETRAC}) and the Unmanned Aerial Vehicle Benchmark (UAVDT)~\cite{du2018unmanned} (mobile camera, 3 possible labels, 80000 images, high density of small objects, see figure~\ref{UAVDT}).

\subsection{Implementations Details}

We implemented the proposed model in Keras~\cite{chollet2015keras} using TensorFlow~\cite{tensorflow2015-whitepaper} as the backend. We used a standard RetinaNet as our baseline, without any bells and whistles or post-processing. We want to keep the models simple in order to properly show the contributions of our architecture and fusion module.

We built a feature pyramid with five different levels, called P3, P4, P5, P6, P7, with the outputs of block 3, 4, 5 of VGG-16. P3 to P5 are the pyramid levels corresponding to block 3 to 5. P6 and P7 are obtained via convolution and down-sampling of P5, and their size is reduced in half at each level: P6 is the half the size of P5, and P7 is the half the size of P6. This is standard for RetinaNet. 

For UAVDT and UA-DETRAC, we adapted the scales used for the anchor boxes by reducing them, due to the high number of small objects in the tested datasets. Instead of using the classic $2^0, 2^{(1.0 / 3.0)}, 2^{(2.0 / 3.0)}$ scale ratios, we used $2^0$, $1 / (2^{(1.0 / 3.0)})$, $1 / (2^{(2.0 / 3.0)})$. This modification did not affect the results on UA-DETRAC, but improved them on UAVDT, causing a bigger gap with the reported state-of-the-art results in the paper. Since we use the same scales for our baseline, this has no effect on our conclusions. The focal loss parameter $\gamma$ is 2 and we used an initial learning rate of 1e-5.

To train both the model and the baseline, we used the adam optimizer~\cite{kingma2014adam}. In order to fit the model into memory, we had to freeze the first four convolutional blocks of the VGG-16 model during training, and only retrained the other weights, with a batch size of one. For a fair comparison, we used the same training setting for our baseline. Despite this limitation, we still achieve state-of-the-art results when compared to single frame object detectors. Even though the first four convolutional blocks are frozen, they are still initialized with fine-tuned weights for each dataset. Note that the weights used to initialize the backbone are the same for the baseline and the model. 

To select the hyperparameter $n$ of our method (the number of frames used before and after), we used a validation set and tried a few values. $n=2$ was the value that worked best for us, so that is the value we use for the final results. We show the results of different values of $n$ in an ablation study in table~\ref{results-ablation}.

\subsection{Performance Evaluation}
For the two datasets, the test set is predetermined and cannot be used for training or to fix hyperparameters. We split the training data into training and validation by choosing a few whole sequences for validation, and the others for training. We did this to prevent overfitting on the validation data that would likely happen if we would split randomly between all frames. We trained the models until the validation loss started to increase, meaning the model was overfitting.

The performance measure used for evaluation is the mAP, meaning Mean Average Precision. The mAP is the mean AP for every class. The AP is the average precision considering the recall and precision curves; thus, it is the area under the precision-recall curve. The minimum intersection over union (IOU) between the ground-truth and the prediction bounding box, to consider a detection valid, is 0.7 for UA-DETRAC and UAVDT, as defined by the dataset’s protocols. The IOU, or the Jaccard index, is the intersection area between two rectangles divided by the union area between them.

\subsection{Results}

\subsubsection{Results on UA-DETRAC}
Results on the UA-DETRAC dataset are reported in table~\ref{resultsuadetrac}. We drew the ROC curves for our model, the baseline and few other state-of-the-art models in figure~\ref{rocua}. Our detector outperforms all classic state-of-the-art models evaluated on UA-DETRAC as well as the baseline by a significant margin.

Something interesting to notice is that our model outperforms R-FCN for the categories labeled ``hard'' and ``cloudy'', confirming our hypothesis that the features are indeed enhanced for hard cases like occlusion and blur (from motion or from clouds). As a result, it raised the mAP for ``overall'' above R-FCN's ``overall''. We have to keep in mind that most VGG-16 layers are frozen during training, and that the final score would probably be much higher if this was not case. Nonetheless, our model convincingly surpasses the baseline in all categories, showing that features are enhanced not only for hard cases, but at all times. We outperform other video object detection for which we found results on UA-DETRAC, that is, 3D-DETNet~\cite{3D_detnet_li20183d} and RN-D-from-scratch~\cite{perreault2019road}. The other video object detectors mentioned in the related works section did not produce results on this dataset.

\begin{table}[t]
\footnotesize
\setlength\tabcolsep{3pt} 
\def\arraystretch{1.5}
\centering
\caption{mAP reported on the UA-DETRAC test set compared to our baseline as well as classic state-of-the-art detectors. Results for ``Ours'' and ``RN-VGG16'' are generated using the evaluation server on the UA-DETRAC website, 3D-DETNet~\cite{3D_detnet_li20183d} is reported as in their paper, and others are as reported in the results section of the UA-DETRAC website. \textbf{Boldface}: best result, \textit{Italic}: baseline.}
\label{resultsuadetrac}
\begin{tabular}{c|c|c|c|c|c|c|c|c}
Model & Overall & Easy & Medium & Hard & Cloudy & Night & Rainy & Sunny \\
\hline
\hline
RN-VID (Ours) & \textbf{70.57}\% & 87.50\% & 75.53\% & \textbf{58.04}\% & \textbf{80.69}\% & 69.56\% & 56.15\% & 83.60\% \\
\hline 
R-FCN~\cite{RFCN_NIPS2016_6465} & 69.87\% &	\textbf{93.32}\% &	\textbf{75.67}\% &	54.31\% &	74.38\% &	\textbf{75.09}\% &	\textbf{56.21}\% &	\textbf{84.08}\% \\
\hline 
\textit{RN-VGG16} & 69.14\% & 86.82\% & 73.70\% & 56.74\% & 79.88\% & 66.57\% & 55.21\% & 82.09\% \\
\hline
EB~\cite{EB_wang2017evolving} & 67.96\%	& 89.65\% &	73.12\% & 53.64\% & 72.42\% & 73.93\% & 53.40\% & 83.73\% \\
\hline 
Faster R-CNN~\cite{ren2015faster} & 58.45\% &	82.75\% &	63.05\% &	44.25\% &	66.29\% &	69.85\% &	45.16\% &	62.34\% \\
\hline 
YOLOv2~\cite{Redmon_2017_CVPR_YOLO2} & 57.72\% &	83.28\% &	62.25\% &	42.44\% &	57.97\% &	64.53\% &	47.84\% &	69.75\% \\
\hline 
RN-D~\cite{perreault2019road} & 54.69\% &	80.98\% &	59.13\% &	39.23\% &	59.88\% &	54.62\% &	41.11\% &	77.53\% \\
\hline 
3D-DETnet~\cite{3D_detnet_li20183d} & 53.30\% &	66.66\% &	59.26\% &	43.22\% &	63.30\% &	52.90\% &	44.27\% &	71.26\% \\
\end{tabular}
\vspace{-4mm}
\end{table}

\subsubsection{Results on UAVDT}
\begin{figure}[t]%
    \centering
    \subfloat[]{\includegraphics[width=0.4\linewidth]{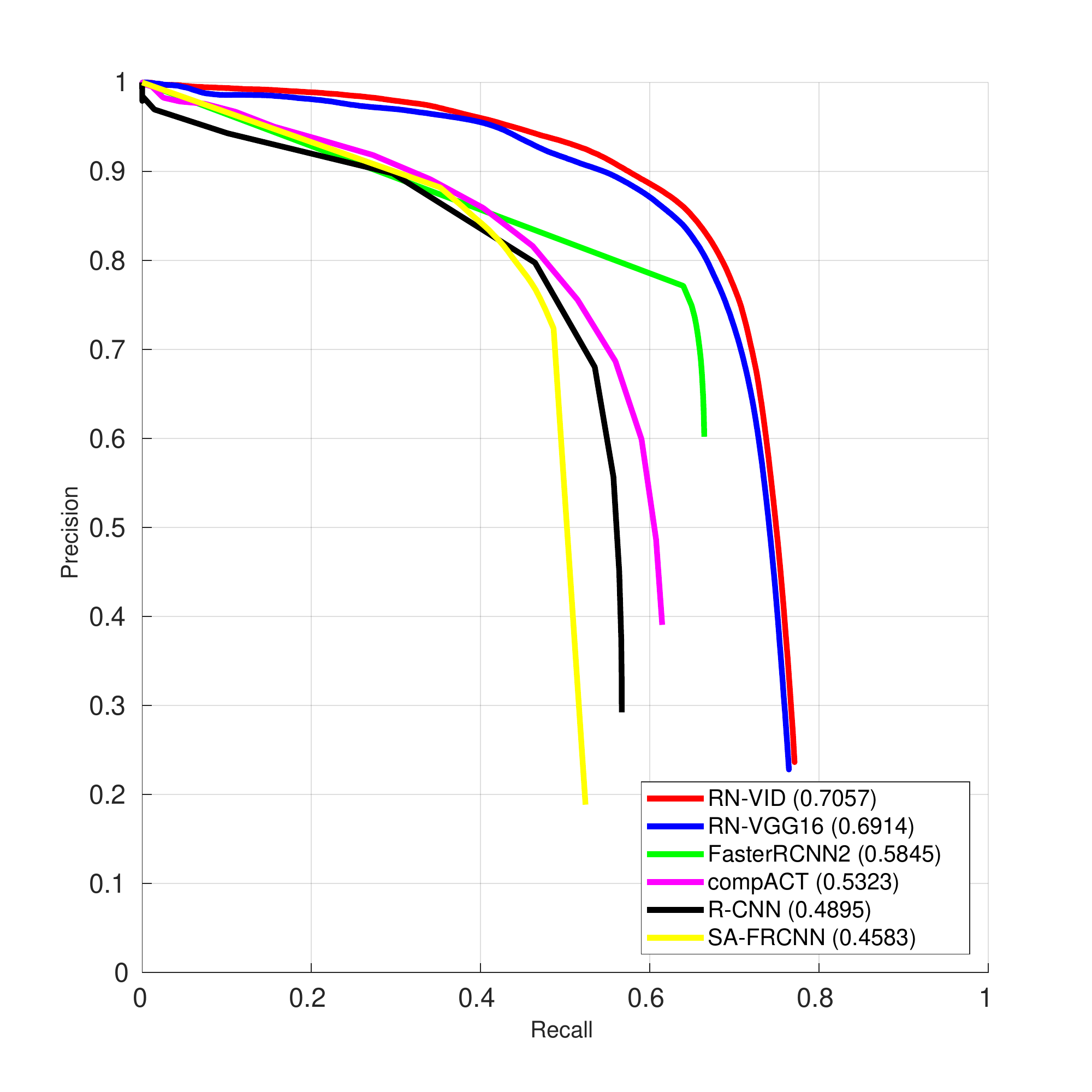}\label{rocua}}
    \hspace{2.5em}
    \subfloat[]{\includegraphics[width=0.4\linewidth]{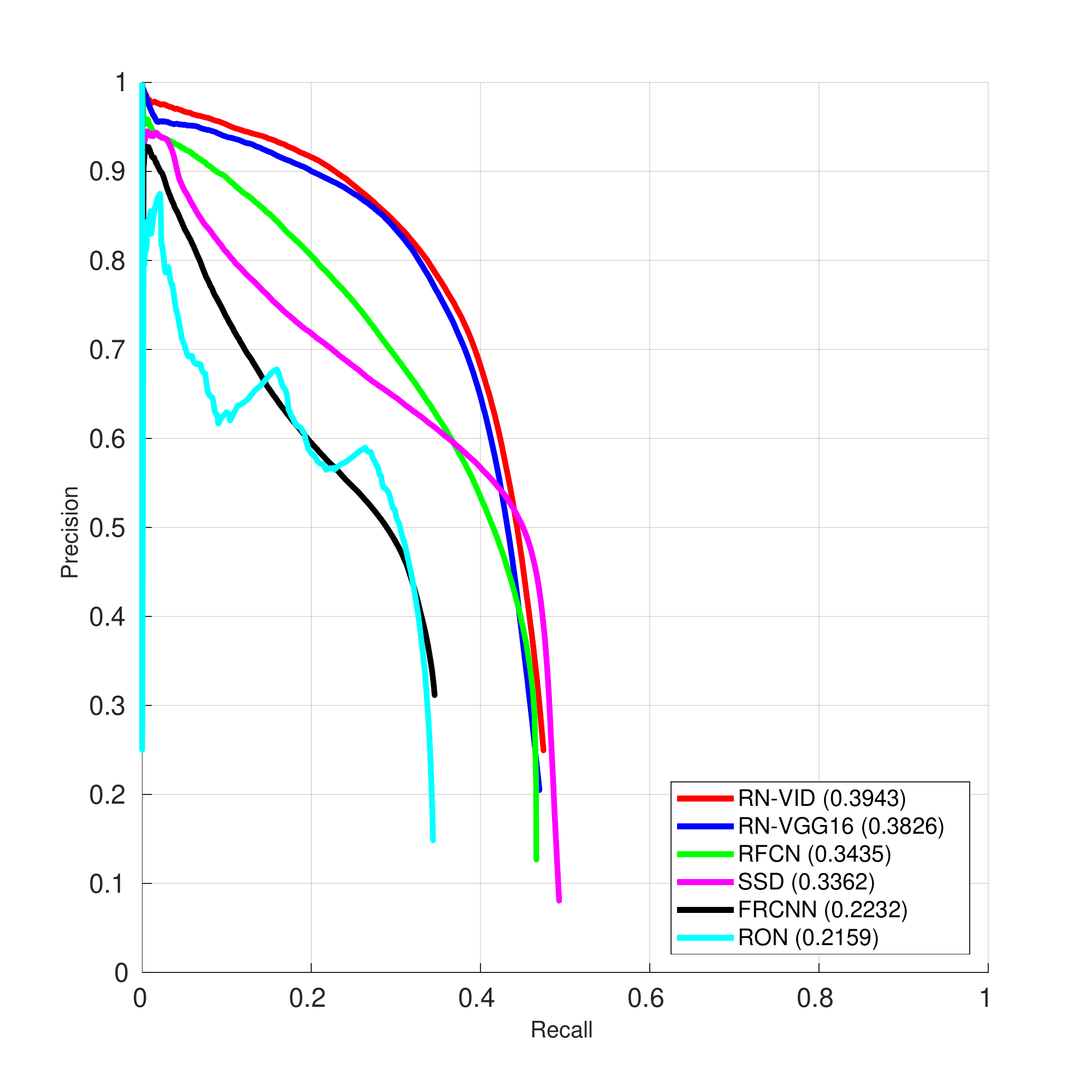}\label{rocUAVDT}}
    \vspace{-0.5em}
    \caption{Precision-Recall curves on UA-DETRAC~\cite{Wen2015Tracking} (a) and UAVDT~\cite{du2018unmanned} (b) for RN-VID (Ours), RN-VGG16 (Baseline) and a few other state-of-the-art methods.}
\end{figure}

The results on UAVDT dataset are reported in table~\ref{results-UAVDT}. We drew the ROC curves for our model, the baseline and few other state-of-the-art models in figure~\ref{rocUAVDT}. Our detector outperforms all classic state-of-the-art models evaluated on UAVDT as well as the baseline by a significant margin. The mAP scores on this dataset are quite low compared to UA-DETRAC due to its very challenging lighting conditions, weather conditions and smaller vehicles. We show that by adapting the scales used for the anchor boxes on each dataset, we can greatly improve results. Also, our model shows results on UAVDT that are consistent with UA-DETRAC's results, having an improvement of $\sim$1.2 mAP points against the $\sim$1.4 on UA-DETRAC.   

\begin{table}[t]
\footnotesize
\setlength\tabcolsep{3pt} 
\def\arraystretch{1.5}
\centering
\caption{mAP reported on the UAVDT test set compared to our baseline as well as classic state-of-the-art detectors. Results for "Ours" and "RN-VGG16" are generated using the official Matlab toolbox provided by the authors, others are reported as in their paper. \textbf{Boldface}: best result, \textit{Italic}: baseline.}
\label{results-UAVDT}
\begin{tabular}{c|c}
Model & Overall \\
\hline
\hline
RN-VID (Ours)& \textbf{39.43}\%\\
\hline
\textit{RN-VGG16}& 38.26\%\\
\hline
R-FCN~\cite{RFCN_NIPS2016_6465} &34.35\%\\
\hline
SSD~\cite{liu2016ssd} & 33.62\%\\
\hline
Faster-RCNN~\cite{ren2015faster} & 22.32\%\\
\hline
RON~\cite{kong2017ron} & 21.59\%\\
\end{tabular}
\vspace{-4mm}
\end{table}

\section{Discussion}
To explain the gains we get from our model, we now discuss a few reasons why aggregating features from adjacent frames is beneficial. 

\subsection{Analysis}
\textbf{Small objects:} The smaller the object, the harder it will be to detect and classify, as a general rule. There is a large number of small objects in the evaluated datasets, as there is in traffic surveillance scenes in general. Having multiple frames allows RN-VID to see the object from slightly different angles and lighting conditions, and a trained network can combine these frames to obtain richer features.

\textbf{Blur:} Blur is omnipresent in traffic surveillance datasets due to road users’ constant motion (motion blur), and weather/lighting conditions. A blurred object can be harder to classify and detect. Since its appearance is changed, the network could recognize it as none of the predetermined labels, and not considering it as a relevant object. Having multiple slightly different instances of theses objects allows the network to refine the features and output finer features to the classification sub-network, finer than each single frame separately. It could also simply choose the best frame is that seems useful. A convincing example of our model performing better in blurry conditions is the ``Cloudy'' category in which it got the best result.

\textbf{Occlusion:} Occlusion from other road users or from various road structures is very frequent in traffic surveillance scenes. Having access to adjacent frames gives our model a strong advantage against temporary occlusions, allowing it to select features from less occluded previous or future frames, making the detections more temporally stable. Figure~\ref{challenges} shows a qualitative example of our model performing better than the baseline in a case of occlusion.   

\subsection{Ablation Study}

To properly assess the contribution of each part of our model, we performed an ablation study. We tried to isolate, as best as we could, our two contributions and looked at the impact of each of them. We justify the choice of using five consecutive frames with an experiment in which we varied this parameter on the UAVDT dataset. We tried several combinations and reported results in table~\ref{results-ablation}. We can see than using five frames is better than using three, and that using three is better than using only one. We did not test with seven frames due to GPU memory issues.

To remove the contribution of the fusion module, we trained a model where instead of merging the feature maps, we would simply concatenate them and continue to build the feature pyramid as usual, by adjusting the kernel size of the convolutions to adapt to the new input size. Doing this actually degrades the performance a lot as shown by the RN-VID-NO-FUSION model in table~\ref{results-ablation}. This is easily understandable by the fact that combining feature maps like this is noisy, and we might need much more data and parameters in order to make this work. We can conclude from this that the fusion module is an essential part of our model.

\begin{table}[t]
\footnotesize
\setlength\tabcolsep{3pt} 
\def\arraystretch{1.5}
\centering
\caption{mAP reported on the UAVDT test set for different variations of our model to conduct an ablation study. Results are generated using the official Matlab toolbox provided by the authors. Number of frames is the number of frames used for each detection.}
\label{results-ablation}
\begin{tabular}{c|c|c}
Model & num. frames& Overall \\
\hline
\hline
RN-VID (Ours) & 5 & \textbf{39.43}\%\\
\hline
RN-VID & 3 & 39.05\%\\
\hline
RN-VGG16 (Baseline) & 1 & 38.26\%\\
\hline
RN-VID-NO-FUSION & 5 & 26.95\%\\
\end{tabular}
\vspace{-4mm}
\end{table}

\subsection{Limitations of our Model}
A limitation of our model is for border situations, the first and last frames of a sequence where we cannot use our architecture to its full potential. However, this is not a problem since we can do a padding by repeating the first and last frame the number of times needed to without a real loss of performance. Also, it takes more memory to train the model then its single frame counterpart, due to the fact that we need multiple frames to train one single ground-truth example.

\section{Conclusion}
A novel approach for video object detection named RN-VID was introduced, composed of an object detection architecture and a fusion module for merging feature maps. This model was trained and evaluated on two different traffic surveillance datasets, and compared with a baseline RetinaNet model and several classic state-of-the-art object detectors. We show that by using adjacent frames, we can increase mAP by a significant margin by addressing challenges in the traffic surveillance domain like occlusion, motion and general blur, small objects and difficult weather conditions.  

\vspace{-0.25em}

\bigskip\noindent\textbf{Acknowledgments.} We  acknowledge  the  support  of  the  Natural  Sciences and  Engineering  Research  Council  of  Canada  (NSERC), [RDCPJ 508883 - 17], and the support of Genetec. The authors would like to thank Paule Brodeur for insightful discussions.

\vspace{-1em}

{
\bibliographystyle{splncs04}
\bibliography{bib}
}
\end{document}